\documentclass[sigconf]{acmart}
\newcommand{\methodname}{ATM~}
\newcommand{\ie}{\emph{i.e.,~}}
\newcommand{\eg}{\emph{e.g.,~}}
\newcommand{\wrt}{\emph{w.r.t.~}}
\newcommand{\etc}{\emph{etc.~}}

\usepackage{soul}
\usepackage{graphicx}
\usepackage{multirow}

\AtBeginDocument{%
  \providecommand\BibTeX{{%
    \normalfont B\kern-0.5em{\scshape i\kern-0.25em b}\kern-0.8em\TeX}}}

\copyrightyear{2023} 
\acmYear{2023} 
\setcopyright{acmlicensed}\acmConference[MM '23]{Proceedings of the 31st ACM International Conference on Multimedia}{October 29-November 3, 2023}{Ottawa, ON, Canada}
\acmBooktitle{Proceedings of the 31st ACM International Conference on Multimedia (MM '23), October 29-November 3, 2023, Ottawa, ON, Canada}
\acmPrice{15.00}
\acmDOI{10.1145/3581783.3612509}
\acmISBN{979-8-4007-0108-5/23/10}

\begin{document}

\title{ATM: Action Temporality Modeling for Video Question Answering}

\author{Junwen Chen}
\email{chenjunw@msu.edu}
\orcid{0000-0001-9808-1520}
\affiliation{%
  \institution{Michigan State University}
}

\author{Jie Zhu}
\email{zhujie4@msu.edu}
\orcid{0009-0004-6119-1063}
\affiliation{%
  \institution{Michigan State University}
}

\author{Yu Kong}
  \email{yukong@msu.edu}
  \orcid{0000-0002-4441-0648}
\affiliation{%
  \institution{Michigan State University}
}






\renewcommand{\shortauthors}{Junwen Chen, Jie Zhu, \& Yu Kong}

\begin{abstract}
Despite significant progress in video question answering (VideoQA), existing methods fall short of questions that require causal/temporal reasoning across frames. This can be attributed to imprecise motion representations.   
We introduce Action Temporality Modeling (ATM) for temporality reasoning via three-fold uniqueness: (1) rethinking the optical flow and realizing that optical flow is effective in capturing the long horizon temporality reasoning; (2) training the visual-text embedding by contrastive learning in an action-centric manner, leading to better action representations in both vision and text modalities; and (3) preventing the model from answering the question given the shuffled video in the fine-tuning stage, to avoid spurious correlation between appearance and motion and hence ensure faithful temporality reasoning. 
In the experiments, we show that \methodname outperforms previous approaches in terms of the accuracy on multiple VideoQAs and exhibits better true temporality reasoning ability.
\end{abstract}

\ccsdesc[500]{Computational methodologies-Artificial intelligence-Computer Vision}

\keywords{Video Question Answering, Action, Static Bias}


\maketitle

\section{Introduction}
\label{sec:intro}

Video question answering (VideoQA) is an interactive AI task, which enables many downstream applications e.g. vision-language navigation and communication systems. It aims to answer the natural language question given the video content. 
Recent VideoQA benchmark~\cite{xiao2021next} has gone beyond the understanding of descriptive content like ``\textit{A baby is crying}'' and started to provide effective diagnostics for the models on solving temporal reasoning and causal reflection, \eg ``\textit{The train stops after moving for a while}''. To accurately answer the question, a VideoQA model needs to detect the object ``train'', recognize the ``railway'' scene, more importantly, ground the action ``move'' and ``stop'' and understand their temporal relations. The questions are unconstrained and complex, and thus, it is necessary to have a visual-text model that has the reasoning capability toward all aforementioned contents.

\begin{figure}[t]
\centering
\includegraphics[width=0.85\linewidth]{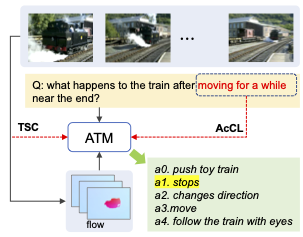}
  \caption{ATM addresses VideoQA featuring temporal reasoning by (1) an appearance-free stream \ie optical flow to extract precise motion cues, (2) action-centric contrastive learning (AcCL) for action-plentiful cross-modal representation, and (3) a temporal sensitivity-aware confusion (TSC) loss to avoid learning a shortcut between temporality-critical motion and appearance. 
  }
\label{fig:teaser}
\end{figure}
Recent advanced VideoQA models have shown the capability of learning from the descriptive contents~\cite{lei2018tvqa, lei2019tvqa+}, thanks to the success of cross-modal transformers~\cite{li2020hero,lei2021less}. However, the temporality reasoning in videos remains a great challenge, since these VideoQAs are only capable of holistic recognition of static content in a video. 
Recent work attempts to solve this issue by (1) enhancing the video representation with fine-grained dynamics~\cite{xiao2022vgt,xiao2022hqga} and (2) answering by grounding to question-critical visual evidence~\cite{li2022equivariant,li2022invariant}. 
But it is hard to achieve a precise grounding, without the ground-truth of temporal boundaries for training. 
The state-of-the-art method VGT ~\cite{xiao2022vgt} proposes to model the atomic actions across frames from the spatio-temporal dynamics of objects. 
In this way, the fine-grained dynamics can be captured. But their model may 
rely on the static bias \ie object appearance, as shortcuts from videos while the causal factors \ie the dynamics are overlooked in training. 
In this paper, we address the importance of precise and faithful modeling of actions for the VideoQA task.

We propose Action Temporality Modeling (ATM) to address the challenging temporality VideoQA (as shown in Fig.~\ref{fig:teaser}). 
A promise of VideoQA compared to ImageQA is to examine the temporal relation reasoning regarding motion information. As the targeted video is continuous, actions across a long video usually share the same scene in short moments. 
We realize that (1) leveraging an appearance-free stream \eg optical flow as input, though the flow stream may become less considered in recent action recognition methods~\cite{bertasius2021space,feichtenhofer2019slowfast}, is still important in VideoQA. Because flow can capture the subtle transition in long horizon and aid the temporality reasoning. 
(2) ATM trains the visual-text encoding in a contrastive manner. Questions are usually unconstrained in the real world. Action may be only a small portion of the question, which is easily overwhelmed by other information such as objects. 
To learn an action-plentiful cross-modal embedding, we develop a novel action-centric contrastive learning (AcCL) before fine-tuning VideoQA. 
Specifically, it parses an action phrase from a question and encourages a feature alignment between the video and the parsed action phrase alone, discarding other textual information. 
The merit of the AcCL is that both video and text encoders are trained to focus on actions, mitigating the backbone's representation bias towards the static visual appearance in videos. 

Based on the learned representations, we further introduce a novel temporal sensitivity-aware confusion loss (TSC) in VideoQA finetuning.
It prevents a model from answering a temporality question if the corresponding video is shuffled in the temporal domain, thus avoiding simply learning the shortcut correlation to the static content.
Note that VideoQA contains a lot of descriptive questions that can be answered invariant to temporal change. Thus, we 
only apply the confusion loss to temporal-sensitive questions that contain temporal keywords.

Thanks to these components, the proposed \methodname outperforms all of the existing methods on three commonly used VideoQA datasets. It is worth noting that our method without external vision-language pretraining can surpass the existing method that relies on large-scale pre-training by a clear margin.
Moreover, we devise a new metric that quantifies the accuracy difference between conditioned on a full video and conditioned on a single frame, which reveals the VideoQA's true temporality reasoning ability.
Results show that our model experiences a larger performance escalation from a single frame to a full video, which demonstrates ours relies on less appearance bias and handles temporal reasoning in a faithful manner.
To summarize, our main contributions are as follows:
\begin{itemize}
\vspace{-1mm}
    \item We propose the \methodname to address VideoQA featuring temporal dynamic reasoning by a faithful action modeling. Our action-centric contrastive learning learns action-aware representations from both vision and text modalities. We realize an appearance-free stream is effective in the multi-event temporality understanding across frames. 
\item  We fine-tune the model with a newly developed temporal sensitivity-aware confusion loss that mitigates static bias in temporality reasoning.
\vspace{-1mm}	
 \item Our method is more accurate than all existing methods on three widely used VideoQA datasets. By a new metric, we also indicate that our method addresses temporality reasoning more faithfully.  
\end{itemize}

\section{Related Work}
\textbf{Video Question Answering}. 
Escalating ImageQA~\cite{antol2015vqa}, VideoQA~\cite{xu2016msr,yu2019activitynet,li2016tgif,xiao2021next,li2022representation,lei2018tvqa} is enriched with reasoning about temporal nature. Prior arts~\cite{le2020hierarchical,park2021bridge,xiao2022hqga} on VideoQA focus on learning an informative video content representation and a cross-modal fusion model to answer the question. An informative video representation is usually hierarchical, fusing object-, frame- and clip-level representations, which are extracted by graph neural network~\cite{jiang2020reasoning, li2022invariant,park2021bridge}, relation learning or transformers. While those VideoQA methods achieve compelling results on VideoQA benchmarks, they mainly answer descriptive questions for the video content, such as questions that holistic recognize the main actions/objects across frames.

\begin{figure*}
    \centering
    \includegraphics[width=0.9\textwidth]{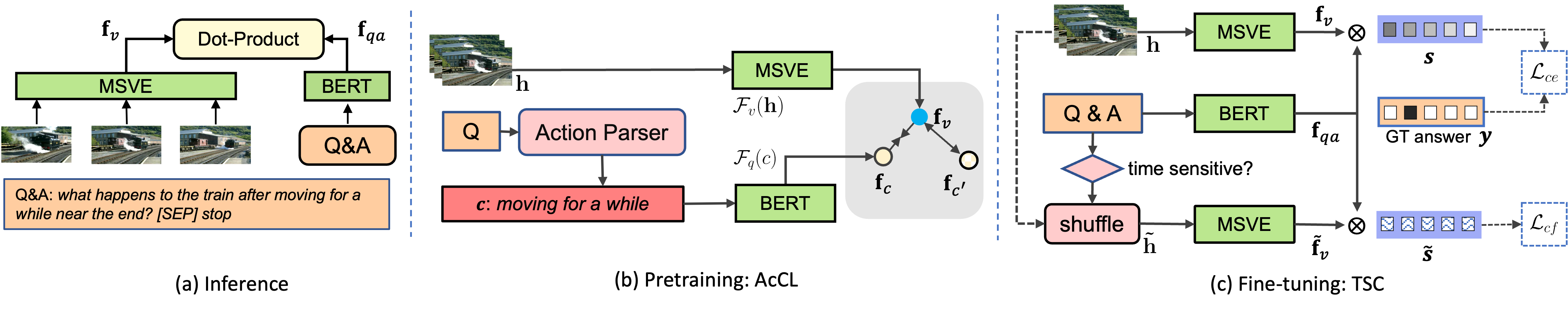}
    \caption{Framework Overview. Following the recent VQAs~\cite{xiao2022vgt, yang2021just}, we solve VideoQA by a similarity comparison between video and text (a). To achieve this, we formulate the training procedure into two stages. Before finetuning, we present a novel action-centric
contrastive learning (AcCL) to guide the visual and text representation expressive for action information (b). After that, we fine-tune the VideoQA (c) by a newly developed temporal sensitivity-aware confusion loss (TSC) to prevent leveraging static bias in temporality reasoning. }
    \label{fig:overview}
\end{figure*}

Recent benchmark~\cite{xiao2021next} begins to challenge the temporal relationship reasoning ability, as actions in videos are diverse and causally dependent. Those methods that are only capable of descriptive content recognition cannot perform well, because they hardly capture the subtle transitions in the same scene in long-horizon. To this end, recent work~\cite{xiao2022hqga, xiao2022vgt} proposes to encode video as a local-to-global dynamic graph of spatiotemporal objects, so that the interaction relations can be encoded. However, the VideoQA models~\cite{xiao2022hqga, xiao2022vgt,gao2023mist} built upon the dynamic graph of patches may easily be distracted by the object's appearance and capture limited motion information. 
We alleviate the distraction by a novel two-stage training to ensure a faithful representation of motions that are critical for temporality reasoning. Specifically, we propose a novel contrastive learning in which the objectives that are the parsed action phrases in questions and a novel confusion loss to prevent question answering if the video is temporally shuffled. 

Concurrent work HiTEA~\cite{ye2022hitea} also introduces temporal shuffling, but the shuffling is used as evaluation test to metric the temporal reliance of datasets and VLP models, while our method leverages this in training a temporal reliable VideoQA. Concurrent work CoVGT~\cite{xiao2023contrastive} also investigates the contrastive learning in VideoQA. But as our action-centric contrastive learning aims at learning a faithful action representation, our contrastive objectives are action phrase in question, different from CoVGT’s question and QA pairs. 
Concurrent work Verbs in Action~\cite{momeni2023verbs} also proposes to improve verb understanding in Video-language task. Verbs in Action focuses on training by the generated the new captions with hard mined verb based on large language model, while our AcCL extracts the action phrase and encourages learning motion representations agnostic to the appearance information.


\textbf{Static Bias in Video}. 
The uniqueness of video lies in the potential to go beyond image-level understanding of the static content \ie scenes, objects, and people to evaluate the temporality reasoning ability of multiple events.
However, for many video(+language) tasks and datasets, given just a single frame of video, an existing image-centric model can achieve surprisingly high performance, comparable to the model using multiple frames. The strong single-frame performance suggests that the video representation is biased towards the static appearance information, namely ``static appearance bias''. 
Existing work~\cite{buch2022revisiting, lei2022revealing,li2018resound,choi2019can} reveals this kind of bias in action recognition dataset~\cite{carreira2017quo,soomro2012ucf101} and retrieval dataset~\cite{liu2019use,luo2021clip4clip}. 
Circling around the fundamental video task action recognition, ~\cite{li2018resound,choi2019can} analyze the role of temporality in action recognition and inspires the subsequent development of profound faithful evaluations~\cite{shao2020finegym,li2018resound} and model structures~\cite{feichtenhofer2019slowfast,lin2019tsm,feichtenhofer2020x3d,duan2022revisiting}. 

To address the challenging temporality reasoning in multi-modal scenarios \ie VideoQA, 
motion representations, unbiased toward appearance, are necessary. 
As VideoQA requires a deep understanding of open-vocabulary action semantics, existing VideoQAs~\cite{le2020hierarchical,xiao2022hqga} extract the motion features based on backbones pre-trained on a large-scale action recognition dataset~\cite{carreira2017quo}. 
As mentioned, static bias exists in action recognition, which makes the motion representations not the causal factors of actions, thus useless to temporality reasoning. Existing methods\cite{choi2019can,li2018resound} mitigate the static bias in action recognition and is evaluated on fine-grained action recognition\cite{shao2020finegym,li2018resound}, where the scene context is the same across the different actions. However, in fine-grained action recognition, motion is the most critical information, which is different from VideoQA where object/entity appearance is inevitable. 

To mitigate static bias in VideoQA, IGV ~\cite{li2022invariant} and EIGV~\cite{li2022equivariant} are proposed to ground the question-critical scenes across frames as the evidence of yielding the answers. 
However, the dominant content of a question is appearance information \eg people, objects, and locations. The grounding may pay less attention to the actions that are critical for temporality understanding and be not precise as no ground truth boundaries are provided. Our method designs two simple yet effective schemes that learn faithful visual and text representations informative for action and temporality. We also revisit the early action recognition work~\cite{wang2016temporal,carreira2017quo} and enhance the motion representation with an appearance-free stream.

\section{Methodology}

Figure \ref{fig:overview} gives an overview of \methodname framework. Our framework addresses the VideoQA task that challenges the temporal reasoning of dynamics in a video. 
Following the recent VQAs~\cite{yang2021just,xiao2022vgt}, we solve VideoQA by a similarity comparison between video and QA pair (Figure \ref{fig:overview}-a). 
To achieve this, we formulate the training procedure into two stages.
In the first stage (Figure \ref{fig:overview}-b), we present a novel action-centric contrastive learning (AcCL, Sec.~\ref{sec:accl}), which makes the visual and text representation expressive for action information. 
After that, we finetune the VideoQA (Figure \ref{fig:overview}-c) by a newly developed temporal sensitivity-aware confusion loss (TSC, Sec.~\ref{sec:tsc}) to prevent leveraging static bias in temporality reasoning. We detailed the video and text encoding in Sec.~\ref{sec:flow}.

\subsection{Preliminaries}
Given a video $\mathbf{h}$ and a question $q$, VideoQA task aims at combining the two modalities $\mathbf{h}$ and $q$ to predict the answer $a$. 
Following existing VideoQA work~\cite{li2022invariant, xiao2022hqga, xiao2022vgt}, we predict the answer by selecting the best matched $a^*$ from many candidates $\mathcal{A}$ of a question $q$, given the corresponding video $\mathbf{h}$:
\begin{equation}\label{Obj}
    a^* = \arg\max\nolimits_{a \in \mathcal{A}} {\mathcal{F}_W(a | q, \mathbf{h}, \mathcal{A})}, 
\end{equation}
where $\mathcal{F}_W$ denotes the mapping function with learnable parameters $W$. The candidates $\mathcal{A}$ are multi-choices in multi-choiceQA or a global answer list in open-ended QA. 

Prior arts on VideoQA usually build $\mathcal{F}_W$ as a cross-attention transformer~\cite{zhu2020actbert, lei2021less}, which takes a holistic token sequence containing video, question and each candidate answer as input and classifies the answers as output.
Recent work VGT~\cite{xiao2022vgt} and VQA-T~\cite{Yang_2021_ICCV} propose to design $\mathcal{F}_W$ as two unimodal transformers that encode video and question-answer pair respectively and compare the visual-text similarity for each answer as output:
\begin{equation}
  s_a = \mathcal{F}_v \left (\mathbf{h}  \right )  \mathcal{F}_q \left ( [q; a]\right )^\top,
\label{sim}
\end{equation}
in which $\mathcal{F}_v$ denotes the video encoder and $\mathcal{F}_v \left (\mathbf{h}  \right ) \in \mathbb{R}^d$ is the video' global feature obtained by mean-pooling the features across $T$ frames. Likewise, $\mathcal{F}_q$ denotes the text encoder and $\mathcal{F}_q \left ( [q; a]\right ) \in \mathbb{R}^d$ is the feature vector of a question-answer pair, where $[;]$ indicates the concatenation of question and answer text. 
The visual-text similarity $s_a$ is obtained via a dot-product of video and text features \wrt the answer $a$. The optimal answer is selected by maximizing the similarity score from the candidate in the pool $\mathcal{A}$:
\begin{equation}
    a^*=\arg \max_{a\in{\mathcal{A}}}(s_a).
\end{equation}

\begin{figure}[t]
    \centering
    \includegraphics[width=0.48\textwidth]{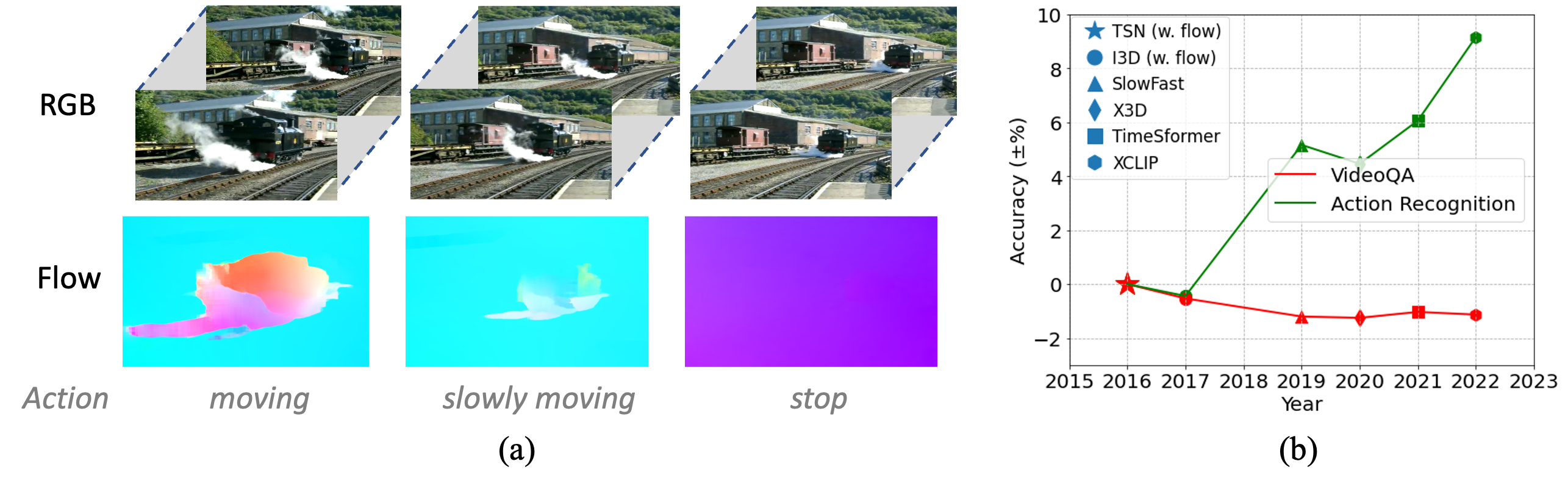}
    \caption{\textbf{Motivation of using an appearance-free stream for motion representation in VideoQA task}. The example in (a) shows the state transition on a train, from moving to stopping. We can see flow provides better cues for the actions than RGB. 
    (b) summarizes the relative performance gain/loss of different video backbones pivot on TSN, for both action recognition (Kinetics~\cite{carreira2017quo}) and VideoQA (NextQA~\cite{xiao2021next}), which shows appearance-free stream \ie flow is necessary for  VideoQA. The numbers for action recognition (green curves) are reported in their paper for Kinetics-400. The numbers for VideoQA are derived based on our implementation on Next-QA.  
    }
    \label{fig:flow}
\end{figure}

Following existing work~\cite{xiao2022vgt, li2022invariant}, we implement $\mathcal{F}_q$ by the BERT~\cite{devlin2018bert} to extract text features. For video modality, many existing methods~\cite{xiao2022vgt, xiao2022hqga} extract features in multiple streams including object-level and frame-level. Following them, we also formulate $\mathcal{F}_v$ as a multi-stream video encoder (MSVE), by which object features are encoded as $f_o \in \mathbb{R}^{{T}\times d}$ and frame features are encoded as $f_r \in \mathbb{R}^{{T}\times d}$. The object/frame feature extraction and transformer-based encoding are exactly the same as state-of-the-art method VGT~\cite{xiao2022vgt} for a fair comparison.

\subsection{Rethinking motion representations in VideoQA}
\label{sec:flow}
In video feature extraction of both the existing methods~\cite{xiao2022vgt, xiao2022hqga, le2020hierarchical} and ours, frame-level features $f_r \in \mathbb{R}^{{T}\times d}$ and object features $f_o \in \mathbb{R}^{{T}\times d}$ both represent appearance. 
Optionally, they~\cite{xiao2022hqga,le2020hierarchical} apply a pre-trained 3D Conv network~\cite{carreira2017quo} on the neighboring frames to capture motions.
However, VideoQA studies the temporality of the actions in a video where multiple actions are performed across frames. As a video captures continuous information, these actions usually share the same scene context and are performed by the same people and on the entity. 
In this case, although 3D Conv can capture motions, neighboring RGB frames may be too redundant to precisely model the actions. For example, in Figure~\ref{fig:flow}-a, it is hard to recognize ``the train is stopping'' in the last clip from RGB. 
This inspires us to enhance the video representation by a stream, where the appearance information is least and hence the motions are highlighted. 
To this end, we resort to optical flow that describes the apparent motion of individual pixels on the image. As shown in  Figure~\ref{fig:flow}-a's example, flow maps provide better cues to understand the state transition of objects \eg ``train'' was moving (in the first and second clip) and stopped (in the third clip).

As VideoQA requires the open-vocabulary semantic understanding of motions, we use the backbone pretrained on a large-scale action recognition dataset Kinetics-400~\cite{kay2017kinetics} to extract flow features.
Flow features are extracted as per appearance frame timestamps as $f_m \in \mathbb{R}^{T \times d}$. To fuse the object, appearance, and flow streams, our MSVE applies MLPs and a learnable multi-head self-attention layer $\text{MSA}$ with position embedding to model the temporal interactions upon the multi-stream features and finally mean-pool the frames to obtain the global video representation $f_v$.
\begin{equation}
f_v = \text{Mean-Pool}(\text{MSA}(\text{MLP}([f_o ; f_r ; f_m])))
\end{equation}
Note that we should not ignore the appearance information in VideoQA task, as the questions are unconstrained and may contain characters, objects and locations that need to be grounded to videos. This is different from the action segmentation~\cite{ding2021temporal} or skeleton-based activity recognition~\cite{zhou2021composer,duan2022revisiting}, where motion is the only critical information.

We revisit the fundamental video understanding task \ie action recognition, in which the early methods \eg TSN~\cite{wang2016temporal, carreira2017quo} also utilized optical flow to capture motions. As shown in Figure~\ref{fig:flow}-b, we observe that although the existing powerful RGB-based backbones e.g. SlowFast~\cite{feichtenhofer2019slowfast}, X3D~\cite{feichtenhofer2020x3d}, TimeSformer~\cite{bertasius2021space} and XCLIP~\cite{ni2022expanding} achieve good performance w/o appearance-free stream i.e. optical flow in action recognition, they are less helpful in VideoQA compared to the early methods w/ appearance-free stream. This demonstrates that towards longer-horizon temporality understanding, a stream free of appearance is necessary. Detailed comparison will be discussed in Sec. \ref{sec:exp_accl}.

\subsection{Action-centric Contrastive Learning (AcCL)}
\label{sec:accl}
As aforementioned, question-answer contains much information including characters, objects, and locations. Actions, the important reasoning objective in videos, may only occupy a small portion of QA text and be neglected in the cross-modal alignment.
Since VideoQA takes the alignment of global video features and a full QA sequence features as the optimization objective, the precise motion information obtained from Sec.~\ref{sec:flow} may not be well exploited.
But a VideoQA model, capable of answering temporal questions, should make good use of motion.

To this end, we propose a novel training scheme that conducts contrastive learning for visual-language matching before finetuning VideoQA objective. Different from conventional VL contrastive learning, the contrastive learning in our method is action-centric. 
It encourages the video representation to be aligned with the representation of \textbf{action phrase} that is parsed from the question. That is to say, other information such as entity, location, objects are not present in the text for matching. For example, in the question ``what happens to the train after moving for a while?'', the action phrase to be aligned with the whole video clip is ``moving for a while''.
Under this matching objective, the video representation has to focus on precise motions, leading to a deep understanding of temporality. 
In specific, we propose a contrastive loss $\mathcal{L}_{pt}$ to update the encoders $\mathbb{F}_v, \mathbb{F}_q$:
\begin{equation}
    \mathcal{L}_{pt}=\sum\limits_i \log\left(\frac{\exp{(s_c)}}{\exp{(s_c)}+\sum\nolimits_{c'\in \mathcal{N}_i} \exp{(s_c')} }\right), 
\label{eq:contrast}
\end{equation}
where $\mathcal{N}_i$ denotes the negative pool of action phrase for the $i$-th sample, i.e., action phrases from the questions that are unpaired to the video $\mathbf{h}$. $s_{c} = \mathcal{F}_v(\mathbf{h})\mathcal{F}_q(c)^{\top}$ is the similarity between the action phrase $c$ and video $\mathbf{h}$ of the $i$-th sample. 
It encourages the video representation closer to its paired action phrase $c$ and far away from the unpaired $c'$ that are randomly sampled into the mini-batch.
Thus, by contrastive to many other action phrases $c'\in{\mathbb{N}_i}$ in the dataset, the motion in vision and the textual action are better mined and aligned. The motion-plentiful features and model provide a good starting point for VideoQA finetuning.

Many VideoQA task benefits from contrastive learning based video language pretraining~\cite{lei2021less,zellers2021merlot} from large-scale video-language data~\cite{bain2021frozen}, which is also reflected in the SoTA model in our task~\cite{xiao2022vgt}. However, our AcCL is just conducted on our task datasets themselves, without resorting to any of the external training data, and has already been more effective than VGT~\cite{xiao2022vgt} with external data pretraining, while taking much less training resources.

\subsection{Temporal Sensitivity-aware Confusion Loss}
\label{sec:tsc}
At the end of Sec.~\ref{sec:flow}, we mention that although we have an appearance-free stream to extract precise motions, the appearance stream is indispensable.
Unfortunately, the appearance stream, even fused with an appearance-free stream, provides the possibility to model action biased towards scene/object context~\cite{choi2019can}.
To mitigate this issue, we propose to prevent the model from answering a question if the corresponding video is randomly ordered in the temporal domain.
Our motivation is that the temporality reasoning requires the model to infer the inter-action relations across temporal, such as ``stop (action 1) \textbf{after} moving for a while (action 2)''. Thus, if the video is randomly shuffled, the ``after'' relation no longer exists. In this case, a reliable network should be unable to answer the ``stop'' to the question like ``What is the train doing after moving for a while?''.

Motivated by this, we design a confusion loss that takes as input the shuffled video $\tilde{\mathbf{h}}$ and question-answer $[q;a]$:
\begin{equation}
\begin{split}
    & \mathcal{L}^{(n)}_{cf} (\hat{\mathbf{p}},\hat{\mathbf{p}}) = -\sum_{j=1}^{|\mathcal{A}|}\hat{p}^{(j)}\log\hat{p}^{(j)}, \\
    & \hat{p}^{(j)} = \frac{\exp{(\hat{s}^{(j)}})}{\sum_{k=1}^{|\mathcal{A}|} \exp{(\hat{s}^{(k)})}}, \\
\end{split}
\end{equation}
where $\hat{s}^{(j)}=\tilde{\mathbf{f}}_{v} \mathbf{f}_q^\top$ (denote the $\hat{s}^{(j)} \in [\hat{s}^{(1)},\ldots,\hat{s}^{(|\mathcal{A}|)}]^\top$) is the inner-product similarity score for the $j$-th answer features $\mathbf{f}_q=\mathcal{F}_q \left ( [q; a]\right )$ \wrt its shuffled video feature vector $\tilde{\mathbf{f}}_{v}=\mathcal{F}_{v} (\mathbf{\tilde{h}})$.
The confusion loss is applied to encourage the maximization of the entropy of the predicted answer distributions over the multiple choices, given the shuffled video. This guides the model to produce confusing classification,
so that the scene context invariant to temporal order change will be ignored in action relation modeling.

Many questions \eg ``Where is the video taken?'' simply rely on descriptive content and can be answered even with the shuffled videos. 
Thus, the confusion loss is only applied to temporal-sensitive questions, \eg the ``after'' question: ``what does A do after raising her hand?''
The temporal-sensitive questions contain specific English syntax, e.g. “before”, “after”, “when”. We filter out the temporal-insensitive questions based on the existence of the syntaxes.
The overall optimization objective is as follows.
\begin{equation}
    \min \mathbb{E}_{q^{(n)}\sim Q_{\tau}}\left[\mathcal{L}^{(n)}_{ce}(\mathbf{y},\mathbf{p}) - \mathcal{L}^{(n)}_{cf}(\hat{\mathbf{p}},\hat{\mathbf{p}})\right],
\end{equation}
where $Q_{\tau}$ denotes the set of questions that are temporally sensitive.  $\mathcal{L}^{(n)}_{ce}$ is the cross entropy loss to metric if the probability over the candidates answers is $\mathbf{p}=[p^{(1)}, ..., p^{(|\mathcal{A}|)}]$ follows ground-truth answer $y$.
$\mathcal{L}^{(n)}_{ce}$ is applied to all of the samples including the temporal-insensitive one, which is to optimize:
\begin{equation}
    \min \mathbb{E}_{q^{(n)}\sim Q_{\setminus{{\tau}}}}\left[\mathcal{L}^{(n)}_{ce}(\mathbf{y},\mathbf{p})\right]
\end{equation}
where $Q_{\setminus{{\tau}}}$ denotes the set of remaining temporally insensitive samples.
The two loss are used for fine-tune the VideoQA after AcCL (see Sec.~\ref{sec:accl}).

\section{Experiments}

\setlength{\tabcolsep}{9pt}
\begin{table*}[t]
\begin{center}
\scalebox{0.8}{
\begin{tabular}{c|ccc|c|ccc|c}
\toprule
\multirow{2}{*}{Methods} & \multicolumn{4}{c|}{NExT-QA Val~} & \multicolumn{4}{c}{NExT-QA Test}  \\ 
\cline{2-9}
                         & Acc@C & Acc@T & Acc@D & Acc@All   & Acc@C & Acc@T & Acc@D & Acc@All   \\
\hline
EVQA \cite{antol2015vqa}  & 42.46 & 46.34 & 45.82 & 44.24 & 43.27 & 46.93 & 45.62 & 44.92 \cr
STVQA \cite{jang2017tgif} & 44.76 & 49.26 & 55.86 & 47.94 & 45.51 & 47.57 & 54.59 & 47.64 \cr
CoMem \cite{gao2018motion}  & 45.22 & 49.07 & 55.34 & 48.04 & 45.85 & 50.02 & 54.38 & 48.54\cr
HCRN* \cite{le2020hierarchical}  & 45.91 & 49.26 & 53.67 & 48.20 & 47.07 & 49.27 & 54.02 & 48.89 \cr
HME \cite{fan2019heterogeneous}  & 46.18 & 48.20 & 58.30 & 48.72 & 46.76 &  48.89 & 57.37 & 49.16 \cr
HGA \cite{jiang2020reasoning}  & 46.26 & 50.74 & 59.33 & 49.74 & 48.13 & 49.08 & 57.79 & 50.01 \cr
HQGA \cite{xiao2022hqga} & 48.48 & 51.24 & 61.65 & 51.42 & 49.04 & 52.28 & 59.43 & 51.75 \cr
P3D-G \cite{cherian2022}  & 51.33 & 52.30 & 62.58 & 53.40 & - & - & - & - \cr
IGV \cite{li2022invariant} & - & - & - & - & 48.56 & 51.67 & 59.64 & 51.34 \cr
EIGV \cite{li2022equivariant} & - & - & - & - & - & - & - & 53.70 \cr 
ATP\cite{buch2022revisiting} & 53.1 & 50.2 & 66.8 & 54.30 & - & - & - & - \cr
VGT \cite{xiao2022vgt} & 52.28 & 55.09 & 64.09 & 55.02 & 51.62 & 51.94 & 63.65 & 53.68 \cr
\textcolor[rgb]{0.5,0.5,0.5}{VGT}* \cite{xiao2022vgt} & \textcolor[rgb]{0.5,0.5,0.5}{53.43} & \textcolor[rgb]{0.5,0.5,0.5}{56.39} & \textcolor[rgb]{0.5,0.5,0.5}{69.50} & \textcolor[rgb]{0.5,0.5,0.5}{56.89} & \textcolor[rgb]{0.5,0.5,0.5}{52.78} & \textcolor[rgb]{0.5,0.5,0.5}{54.54} & \textcolor[rgb]{0.5,0.5,0.5}{67.26} & \textcolor[rgb]{0.5,0.5,0.5}{55.70} \cr
\hline
\bf{Ours}   & \bf{56.04} & \bf{58.44} & \bf{65.38} & \bf{58.27} & \bf{55.31} & \bf{55.55} & \bf{65.34} & \bf{57.03} \cr
\bottomrule
\end{tabular}
}
\caption{Results of multi-choice QA on validation set and test set of NextQA~\cite{xiao2021next} dataset. The \textbf{best} results are bolded. 
Note that the greyed out VGT* uses 0.18 million videos from webvid dataset~\cite{bain2021frozen} as pretraining, while the remaining include \methodname do not pretrain on the external large-scale data. All of numbers for existing work are recorded from their papers. ``-'' indicates the missing results. $Acc_C$, $Acc_T$, $Acc_D$ denote the accuracy for causality, temporality and descriptive questions.
} 
\label{tab:mulresults}
\end{center}
\end{table*}

\subsection{Datasets}

\textbf{NExT-QA} \cite{xiao2021next} consists of 47.7K questions with answers in the form of multiple choices, which is annotated from 5.4K videos. It pinpoints the causal and temporal reasoning over the object interaction. 
Note that the causal questions \eg ``How'', ''Why'' require the corresponding answers visible in the video and thus the causal questions also assess the multi-event temporality understanding.

\textbf{TGIF-QA} \cite{jang2017tgif} contains 134.7K questions about repeated actions, state transitions and a certain frame, which is annotated from 91.8K GIFs. 
\textbf{MSRVTT-QA} \cite{xu2017video} challenges a holistic visual recognition or description, which includes 10K annotated videos and 244K open-ended question-answer pairs.  

\subsection{Implementation Details}

\textbf{Appearance Features}
Following~\cite{xiao2022vgt,xiao2022hqga}, we decode the video into frames and sparsely sample 16 clips where each clip is in the length of 4 frames. 
To make a fair comparison with state-of-the-art VGT~\cite{xiao2022vgt}, we also the RoI aligned features as the object appearance features $f_o \in \mathbb{R}^{16*2048}$ pretrained by ~\cite{anderson2018bottom}.

\textbf{Motion Features}
We use denseflow~\cite{denseflow} to extract the optical flow maps using videos' original FPS. Then, we use mmaction2~\cite{2020mmaction2}-based ResNet from TSN~\cite{wang2016temporal} pre-trained on Kinetics-400~\cite{carreira2017quo} to extract optical flow features. 
We uniformly distribute the flow maps into $K=16$ clips per video and sample 5 frames as per each clip and obtain a $2048$-d feature vector for a clip. Thus, motion features $f_m$ for a video are $\mathbb{R}^{16*2048}$.

\textbf{Action Phrase}
We parse the action phrases from questions using SpaCy parser~\cite{spacy2} Specifically, we use dependency parsing to get the phrases in a question and use the pos-tag to find the verb in the question. Then we filter the phrases that contain the verb and select the shortest one as the action phrase. For example, for the question ``what happens to the train after moving for a while near the end?'', the action phrase is ``moving for a while''.

\textbf{Action-centric Contrastive Learning}
We parse the action phrases from questions using SpaCy parser~\cite{spacy2}.
We use Adam optimizer~\cite{kingma2014adam} with cosine annealing learning schedule of PyTorch initialized at $1e-5$ on NVIDIA RTX A6000 at the maximum epoch of 10 among all of the datasets. 
Each batch contains 64 video-action pairs and forms 64 pairs in total for the contrastive learning.

\textbf{Temporal Sensitivity-aware Confusion Loss}
English questions typically follow a syntactic structure. Temporal-sensitive questions contain specific syntax, \eg ``after'', ``before'', ``... when...", ``...while...'' and \etc. The remaining is descriptive questions, \eg a count question, \textit{``How many people are involved in the video?''}, which is insensitive to time. 
We detect the existence of the syntaxes and filter out the temporal-insensitive questions. 
For Next-QA, we have $17,681$ temporal-sensitive questions and $16,451$ temporal-insensitive questions in the training set.
For T-Gif~\cite{li2016tgif}, as its ``action'' and ``transition'' splits focus on repeated actions and transitions respectively, all of the questions in those splits are temporally sensitive. 
For the open-ended QAs including TGif-FrameQA and MSRVTT~\cite{xu2016msr}, we do not apply the confusion loss as they focus more on the descriptive content. 

Please refer to appendix for additional details.

\subsection{Comparison with State-of-the-Art}
Table ~\ref{tab:mulresults} compares our method with existing state-of-the-art~(SoTA) VideoQA methods on the widely used Next-QA dataset that feature the temporality reasoning. 
To ensure a fair comparison, \methodname follows SoTA VGT~\cite{xiao2022vgt} and uses the exactly same appearance feature extraction and applies DGT~\cite{xiao2022vgt} to model the object features. 
From the table, we can observe that \methodname outperforms all existing methods without external data pretraining, by at least $3.85\%$ and $3.35\%$ on val. and test splits respectively. The outperformance is across causal, temporal, and descriptive splits of the Next-QA dataset, which demonstrate that \methodname is effective in various question types that span from short segment to full video, from causal to temporal, and from single to multiple event execution.

\setlength{\tabcolsep}{1.8pt}
\begin{table}[t!]
    \small
    \centering

    \scalebox{0.9}{
       \begin{tabular}{c|cc|c|cc|c} 
       \toprule
        \multirow{2}{*}{Models} & \multicolumn{5}{c|}{TGIF-QA}               & \multirow{2}{*}{MSRVTT-QA}  \\ 
        \cline{2-6}
                                & Action & Transition &Frame-QA & Action$\dagger$  & Transition$\dagger$ &                             \\ 
        \hline
        LGCN \cite{huang2020location} & 74.3 & 81.1 &56.3 & - & - & -  \\
        HGA \cite{jiang2020reasoning} & 75.4 & 81.0 &55.1 & - & - & 35.5  \\
        HCRN \cite{le2020hierarchical}  & 75.0 & 81.4 &55.9 & 55.7 & 63.9 & 35.6 \\
        B2A \cite{park2021bridge} &75.9 & 82.6 & 57.5 & - & - & 36.9 \\
        HOSTR \cite{dang2021hierarchical}  &75.0 & 83.0 &58.0 & - & - & 35.9 \\
        HAIR \cite{liu2021hair}  &77.8 & 82.3 &60.2 & - & - & 36.9  \\
        MASN \cite{seo2021attend}   &84.4 & 87.4 &59.5 &- & - & 35.2 \\
        PGAT \cite{peng2021progressive}  &80.6 & 85.7 &61.1 & 58.7 & 65.9 & 38.1 \\
        MHN \cite{peng2022multilevel}  &83.5 & 90.8 &58.1 &- & - & 38.6  \\ 
        \textcolor[rgb]{0.5,0.5,0.5}{ClipBERT*} \cite{lei2021less}  &  \textcolor[rgb]{0.5,0.5,0.5}{82.8} & \textcolor[rgb]{0.5,0.5,0.5}{87.8} &\textcolor[rgb]{0.5,0.5,0.5}{60.3} & - & -& \textcolor[rgb]{0.5,0.5,0.5}{37.4}  \\
        \textcolor[rgb]{0.5,0.5,0.5}{SiaSRea*} \cite{yu2021learning}  &  \textcolor[rgb]{0.5,0.5,0.5}{79.7} & \textcolor[rgb]{0.5,0.5,0.5}{85.3} & \textcolor[rgb]{0.5,0.5,0.5}{60.2} & - &- & \textcolor[rgb]{0.5,0.5,0.5}{\underline{41.6}} \\
        \textcolor[rgb]{0.5,0.5,0.5}{MERLOT*} \cite{zellers2021merlot} & \textcolor[rgb]{0.5,0.5,0.5}{94.0} & \textcolor[rgb]{0.5,0.5,0.5}{96.2} & \textcolor[rgb]{0.5,0.5,0.5}{69.5} & - & - & \textcolor[rgb]{0.5,0.5,0.5}{{\bf43.1}}  \\ 
        VGT~\cite{xiao2022vgt} & \underline{95.0} & \bf{97.6} & 61.6 & {\underline{59.9}}  & {\underline{70.5}} & 39.7 \\ 
        \hline
        Ours \cite{zellers2021merlot}  & \bf{96.0}  & \underline{97.3} &61.6 & \bf{65.7} & \bf{71.0} &40.3  \\ 
        \bottomrule 
        \end{tabular}
    }
    \caption{Results on TGIF-QA and MSVTT-QA. $\dagger$ denotes TGIF-QA-R \cite{peng2021progressive} whose multiple choices for repeated action and state transition are more challenging. * denotes the models pretrained with large-scale external data.
    }
    \label{tab:tgif}
\end{table}  

Moreover, \methodname which comes without external large-scale pre-training, even surpasses the existing method that used large-scale pretraining on more than 0.18 million videos~\cite{bain2021frozen}, by a clear margin of $1.38\%$ and $1.33\%$ on validation and test splits respectively.  
This demonstrates that \methodname comprises of appearance-free motion features Sec.~\ref{sec:flow}, action-centric contrastive learning Sec.~\ref{sec:accl} and temporal sensitive-aware confusion objective Sec.~\ref{sec:tsc}, which holistically models action temporality, is more effective than the global video-text matching while uses less training computation resources.

In ATP~\cite{buch2022revisiting}, the temporal modeling is performed on frames that are representative for single events and are encoded with CLIP model~\cite{radford2021learning}. Our method also exceeds ATP~\cite{buch2022revisiting} by a large margin of $3.97\%$. 
This shows in temporality-heavy tasks, precise and faithful motion modeling is more effective than selecting the informative single frame for an event.
This validates that \methodname to precisely model and reason about motion, sets the new SoTA on Next-QA~\cite{xiao2021next} benchmark.

Furthermore, we compare \methodname with SoTA on TGIF-QA in Table~\ref{tab:tgif}. Following the protocol, we use the same appearance features extracted by VGT~\cite{xiao2022vgt} and extract the motion stream features. 
We observe that \methodname set new SoTA for repeated actions, and transition in TGIF-QA, which shows \methodname as a whole is also effective in the repeated action and object transition scenarios. 

For MSRVTT-QA in Table~\ref{tab:tgif}, our performance (free-of pretraining) is better than pretraining-free SoTA VGT but is inferior to the large-scale pre-trained methods MERLOT~\cite{zellers2021merlot} and SiaSRea* \cite{yu2021learning}. 
Our method on TGIF-Frame-QA performs close to the pretraining-free sota VGT~\cite{xiao2022vgt}.
This is because pretraining help model the descriptive content, while our work focuses on action temporality.

\subsection{True Temporality Metric}
ATP~\cite{buch2022revisiting} evaluated the upper bound performance of a single-frame model on a video dataset and pointed out that even though NextQA dataset focuses on temporality reasoning, the dataset still contains static appearance bias. 
A small portion of questions can be correctly answered exclusively from a single frame without temporal information. 
To this end, we propose to measure the temporality faithfulness of VideoQA methods, \ie revealing if a VideoQA method learns true temporality to answering questions, instead of learning the spurious correlation between the static appearance and the answer. 
In specific, the proposed true temporality metric measures the difference of QA accuracy between given the full video and given the middle frame respectively, as $\delta$.
The middle-frame setting is that only the middle clip (7th only among the 16 clips) is taken as vision input for QA, so MSA in Eq. 4 is applied on a single token sequence.

Table~\ref{tab:re_as} shows that \methodname better learns the true temporality compared to SoTA VGT, w/ w/o pretraining, on both Next-QA and TGif-QA. We observe that the external large-scale data for pretraining VGT guides the model to leverage more static information in temporality reasoning (only +0.84\% on Next-QA test) since the pre-training helps more on the descriptive content that is static.
Each of our component \ie AcCL, TSC, and appearance-free motion stream, helps to learn the true temporality. TSC mitigates the static bias by preventing answering temporality question if the temporal relations are destroyed. AcCL encourages learning motion representation agonistic to the entity or other appearance information. Appearance-free motion streams extract motion-plentiful representations that are necessary to understand the true temporality.

\begin{table}[t]
\centering
\scriptsize
\setlength{\tabcolsep}{0.6mm}
\begin{tabular}{l|cccc|cccc}
\hline
  \multirow{2}{*}{} & \multicolumn{4}{c|}{Next-QA (\%)} & \multicolumn{4}{c}{TGIF-QA (\%)} \\
  \cline{2-9}
   & val-Acc  & val-$\delta$ & test-Acc  & test-$\delta$   &act-Acc &act-$\delta$   &trans-Acc &trans-$\delta$ \\ \hline
Ours   & \textbf{58.27} &  +\textbf{5.51} & \textbf{57.03} &  \textbf{+5.13} & \textbf{96.0} & \textbf{+1.2} & 97.3 & \textbf{+1.3}    \\  
w/o AcCL   & 56.87 & +2.71 & 55.02 & +2.30   & 93.5  & +0.5  & 97.1  &+0.7 \\  
w/o TSC    & 57.99  & +2.98 & 56.24  &+3.25  & 95.6 & +0.8 & 96.9  &+0.2   \\

w/o motion stream   & 56.57 & +3.02  &55.78  &+2.80  & 95.2 & +0.9 & 96.7 & +0.8   \\ 
VGT w/o pretrained & 55.02 & +2.91   &53.68  & +2.15  & 95.0 & +0.6& \textbf{97.6} & +0.3 \\  
VGT w/ pretrained  &56.89  & +1.02  & 55.70 & +0.84  &- &- &- &- \\  \hline
\end{tabular}

\caption{True temporality evaluation: Study of model components and comparison with SoTA.}
\label{tab:re_as}
\end{table}

\subsection{Ablation Studies}
In addition to the study of each component's individual contribution, we conduct further ablation studies on NextQA~\cite{xiao2021next} dataset. 

\subsubsection{Impact of Action-centric Contrastive Learning}

We test different variants of the text in Action-centric Contrastive Learning (AcCL). Table~\ref{tab:ghu_bg_fah}-a summarizes the results of the ablations. AcCL aims at learning action features by aligning the video with the action phrase from the question. The variants replace the action phrase by (1) the correct answer text \wrt to video-question, denoted as ``Answer'', (2) the concatenation of the entire question and the correct answer text, denoted as ``Question+Answer'', (3) the entire question text, denoted as ``Question'', (4) the verb in question.

\begin{figure*}[t]
\begin{center}
\includegraphics[width=0.8
\textwidth]{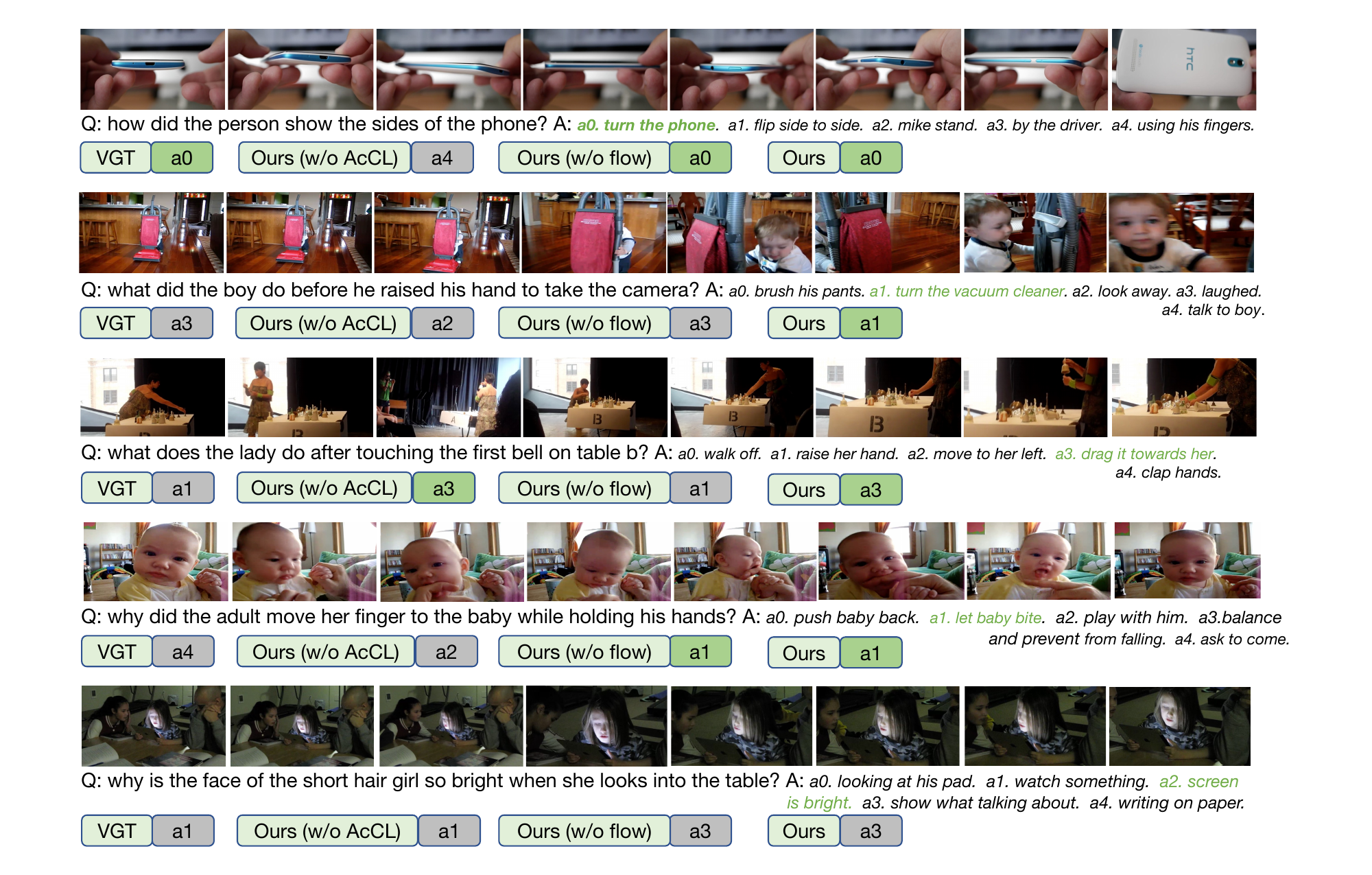}
\end{center}
\caption{{\bf Visualization}. The ground-truth are marked in green. We display the results of \methodname, \methodname w/o AcCL (as ``Ours w/o AcCL''), \methodname w/o optical flow (as ``Ours w/o flow'') and the existing SoTA method VGT~\cite{xiao2022vgt}. The samples span across causality (1,4,5) and temporality (2, 3) reasoning. } 
\label{fig:visualize}
\end{figure*}

\begin{table}[ht]
\centering
\setlength{\tabcolsep}{4pt}
\resizebox{0.5\textwidth}{!}{%
\begin{tabular}{ccc} 
\begin{tabular}{lcc}
\hline
Variants  & val ~(\%) & test~(\%)    \\ \hline
Action phrase (ours)  & \textbf{58.27} & \textbf{57.03}         \\ \hline
Answer        & 55.92 & 53.60  \\
Question+Answer  & 56.83 & 55.16 \ \\ 
Question   & 57.51 & 56.38  \\
Verb in Question   & 57.07  &  56.57    \\ 
w/o AcCL   & 56.87 & 55.02 \\  
\hline

$\:\:\:\:\:\:\:\:\:\:\:\:\:\:\:\:\:\:\:\:\:\:\:\:\:\:\:\:\:\:\:\:\:\:\:\:\:\:\:\:\:$(a) \\ 
\\
\hline
Variants   & test~(\%) & test-$\delta$ ~(\%)   \\ \hline
TS-aware (Ours)   & \textbf{57.03}  & \textbf{+5.13}       \\ \hline
TS-unaware   &  56.89  &+3.67 \\ 
w/o TSC    & 56.24  &+3.25 \\ \hline
\end{tabular}%
&
\begin{tabular}{lcc}
\hline
Variants  & val ~(\%) & test~(\%)   \\ \hline
TSN (ours)   & \textbf{58.27} & \textbf{57.03}       \\  
I3D   & \textbf{57.71} & \textbf{56.40}       \\  \hline    
3D ResNext101   & 57.01  &  55.30 \\ 
SlowFast   & 56.97  &  55.83  \\  
X3D   &  56.27  &  55.78  \\   
Timesformer  & 56.99  &  56.00 \\   
XCLIP & 56.08  &55.90   \\   
I3D-RGB only   & 57.35  &  55.63 \\ 
TSN-RGB only  & 56.94  &  55.42    \\ 
TSN-Flow only    & 56.89 & 55.85          \\ 
I3D-Flow only    & 56.76 & 55.73         \\ 
w/o motion stream   & 56.57 & 55.78      \\ \hline
$K=8$      & 57.63 & 56.66        \\ 
$K=24$    & 57.82 & 56.35       \\ \hline
\end{tabular}

     \\
  (b) & (c) \\
\end{tabular}
}
\caption{Ablation study on different variants of (a) AcCL (b) TSC and (c) motion representations.
}
\label{tab:ghu_bg_fah}
\end{table}
Table~\ref{tab:ghu_bg_fah}-a shows that our AcCL outperforms all of the other variants. We observe that the ``Question'' variant performs $0.65\%$ worse than our ``action in question" on test split since the full question text contains entity, scene, and other appearance information in addition to the action phrase. Contrasting with full questions will distract the representation from the motion information to the dominant and easily learned appearance features, which is less effective than action-centric version. Using ``Answer'', ``Question+Answer'' also performs worse than ours. This demonstrates that the action phrases in questions are the information that the randomly initialized model parameters easily overlook but are important for temporality. 
 Using ``verb from question'' is also less effective, as the action cannot be described by a single word in many cases, e.g. verb ``get'' is not informative enough for the action phrase ``get up''.

\subsubsection{Impact of TSC Loss}
We compare our Temporal Sensitivity-aware Confusion loss (TSC) in Table~\ref{tab:ghu_bg_fah}-b, with variants (1) removing the TSC and only training with cross-entropy, as ``w/o TSC''. (2) applying the confusion loss to all samples regardless of time-sensitivity, as ``TS-unaware''. Our method is slightly better than these two variants in VideoQA accuracy and much higher on the proposed true temporality reasoning metric. This validates that alleviating static bias by TSC helps a faithful temporal reasoning model, which in turn improves the event temporality understanding.

\subsubsection{Impact of Appearance-free stream}
\label{sec:exp_accl}
Table~\ref{tab:ghu_bg_fah}-c shows the ablations on motion features $f_m$ and analyzes the effectiveness of incorporating an appearance-free stream. 
In the table, TSN and I3D extract motion features with an appearance-free stream \ie flow maps, while the remaining extract motions only from the appearance-included input \ie RGB. These RGB-only methods SlowFast~\cite{feichtenhofer2019slowfast}, X3D~\cite{feichtenhofer2020x3d}, TimeSformer~\cite{bertasius2021space} and XCLIP~\cite{ni2022expanding} show superb performance on action recognition, as shown in Fig.~\ref{fig:flow}-b. But they fall behind of the methods with the optical flow on motion representations for VideoQA, though TSN and I3D are relatively early work without fancy network structures. 
RGB frames may be enough for characterizing limited sets of atomic actions that are dominant for action recognition, but it is less effective in modeling events with long-horizon temporality. 
3D ResNext101~\cite{hara3dcnns} has been used for motion feature extraction in existing VideoQA~\cite{le2020hierarchical, xiao2022hqga}, but it is also RGB-only and $1.73\%$ worse than TSN where flow is used. 

In addition, Table~\ref{tab:ghu_bg_fah}-c also shows that the flow maps are helpful when accompanied by the corresponding RGB frames. Motions in VideoQA cannot be extracted purely from an appearance-free stream, since appearance also provides important cues. The table also shows that with the number of clips as per video $K=16$, we achieve the best accuracy which is $57.03\%$ on test split. The accuracy slightly drops if we distributed the videos into clips that are more e.g. $K=24$ or less e.g. $K=8$. This shows that sampling at a certain rate can encode the informative features across multiple visual modalities. But beyond a certain extent of sampling rate, the model may perform worse due to overfitting.

\subsection{Qualitative Analysis}

In Fig.~\ref{fig:visualize}, we qualitatively evaluate the improvement of the \methodname by visualizing the results of representative samples in val. split. We can observe that the AcCL scheme helps to learn the discriminative representations for actions \eg ``turn'' in (1), while the variant w/o AcCL may learn the superficial correlations between appearance \eg ``his fingers'' and the answers. Moreover, the appearance-free stream also helps in extracting precise and useful motions. Since the scene and actor do not change in (3), the optical flow stream is informative for recognizing the ``drag towards'' action. 
We observe that ATM also avoids over-exploiting language bias, as the proposed AcCL helps ground the action text to the visual evidence e.g. ``baby bite'' in (4), while others may rely on the question-answer shortcuts between ``move finger to baby'' and ``a2. play with him''.

ATM focus on action modeling and it may fall short in reasoning about the object characteristics, \eg the ``light screen'' causes the girl in a ``bright face'' in (5).
Large-scale vision-language pre-training augmented with knowledge could be helpful. We leave the knowledge-driven action modeling for future work.

\section{Conclusion}

In this paper, we propose a novel framework to solve the VideoQA featuring temporality reasoning.
To this end, we realize that it is worth revisiting optical flow, as flow may become less considered in atomic action recognition but is still effective in long-horizon temporality. Then, we propose an action-centric contrastive learning that makes both video and text representations informative for action.
Finally, we fine-tune the VideoQA via a novel temporal sensitivity-aware confusion loss to mitigate the potential static bias.
Our ATM method is demonstrated to be superior to all existing VideoQA methods on multiple benchmarks and shows a faithful temporality reasoning via a new metric. 

\textbf{Limitations:}
While \methodname outperforms the existing work, there is ample room for further research. 
Although \methodname can deal with arbitrary-length video, it divides the video into a finite number of clips and extracts features per clip. This may not be adequate to capture enough action information when the action occurs in a very short time over a long duration video. 
Another challenge is time complexity of optical flow computation. It would be worthwhile to study the efficient ways to extract the appearance-free stream. 

~\\[0pt]
\section{Acknowledgements}
Junwen Chen and Yu Kong are supported in part by NSF SaTC award 1949694, and the Office of Naval Research under grant number N00014-23-1-2046. The views and conclusions contained in this document are those of the authors and should not be interpreted as representing the official policies, either expressed or implied, of the Office of Naval Research or the U.S. Government.

\appendix


\end{document}